\documentclass[onecolumn]{tamplate/hexafuture}
\usepackage{environ}
\usepackage{flafter}

\newcommand{\system}{URAI}
\hypersetup{pdftitle={Make Code as Policy Great Again: Frontier Agents Write, Call, and Evolve Robot Tools},
  pdfauthor={Shijia Ge, Alex Zhou, Jianshu Zeng, Yexing Wan, Di Wu, Zelin Zheng, Yazhe Wang, Zhiqi Jia, Xuan Shangguan, Jay Zhu, Yijun Liu, Lingyu He, Sihang Wu, Xiao He, Hongcheng Gao}}

\title{Make Code as Policy Great Again: Frontier Agents Write, Call, and Evolve Robot Tools}

\author{
\begin{center}
    Shijia Ge$^{1}$ \quad Alex Zhou$^{1}$ \quad Jianshu Zeng$^{2}$ \quad Yexing Wan \quad Di Wu$^{3}$
    \\[5pt]
    Zelin Zheng$^{1}$ \quad Yazhe Wang$^{3}$ \quad Zhiqi Jia$^{1}$ \quad Xuan Shangguan \quad Jay Zhu$^{1}$
    \\[5pt]
    Yijun Liu$^{4}$ \quad Lingyu He$^{1}$ \quad Sihang Wu$^{1}$ \quad Xiao He$^{1,\dagger}$ \quad Hongcheng Gao$^{1,4,\dagger}$
    \\[12pt]
    $^{1}$Hexafuture Inc. \qquad $^{2}$Peking University \qquad $^{3}$Beijing Institute of Technology \qquad $^{4}$Tsinghua University
    \\[5pt]
    $^{\dagger}$Corresponding authors
\end{center}
}

\let\hexaabstract\abstract
\let\abstract\relax

\NewEnviron{abstract}{\expandafter\hexaabstract\expandafter{\BODY}\global\let\abstractlist\abstractlist}

\begin{document}
\begin{abstract}
Frontier models can control robots, but reasoning through every reach, grasp, and retreat makes manipulation slow and token-intensive. We revisit code as policy with a different division of labor: models build executable tools, code handles multi-phase motions, and models decide what to do next. We introduce URAI (Universal Robot--Agent Interface), which couples a programming agent that constructs robot tools with an execution agent that uses them in a feedback loop. The programming agent writes reusable and task-specific tools from task intent and refines them through execution feedback and human guidance. The execution agent selects and parameterizes these tools from current observations; each call runs a complete motion locally before returning control to the agent. Unlike delegating subsequent decisions to a generated program, this design retains model-level decision-making between tool executions. Validated tool revisions persist across episodes without updating foundation-model weights, and a shared GUI and API make the same tools available to humans and agents. Across five RoboDojo tasks and four frozen execution agents, URAI raises aggregate success from 18.0\% to 53.0\% relative to direct fingertip control, with the largest gain on Swap Blocks; with the same tools, a program written in advance reaches only 24\% against 56\% for two agents deciding after each call. Three of the four agents also finish episodes 1.3--1.5$\times$ faster with 1.5--1.7$\times$ fewer execution-agent output tokens; DeepSeek-V4-Flash's cost barely changes. We further evaluate URAI on seven real-world AgileX dual-arm tasks, spanning object manipulation, cloth folding, and human-interactive tic-tac-toe. URAI connects the coding and decision-making capabilities of frontier agents, organizing robot control around reusable tools that agents can both invoke and revise.
\end{abstract}

\maketitle
\justifying
\setlength{\parindent}{0pt}
\section{Introduction}
\label{sec:introduction}

\begin{figure}[t]
\centering
\includegraphics[width=\textwidth]{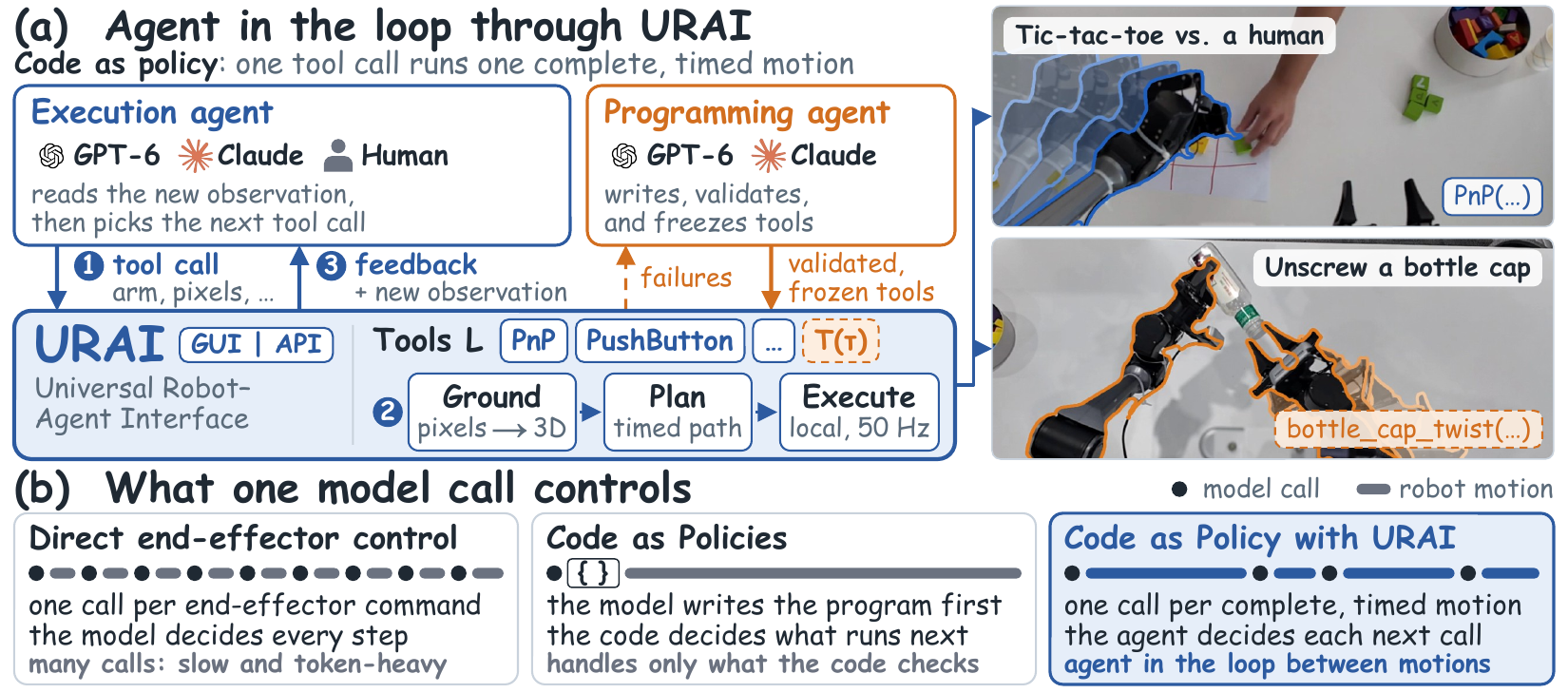}
\caption{\textbf{\system{} overview.} (a) A programming agent writes and validates reusable and task-specific tools. An execution agent selects tools from observations and feedback, while a shared GUI/API backend grounds, checks, and executes multi-phase motions locally. Fading outlines in the demonstration-video frames indicate earlier arm poses within one tool call. (b) Decision boundaries: direct control queries the model for each end-effector command; Code as Policies delegates subsequent decisions to generated code, which may query perception; \system{} returns control to the agent between tool executions.}
\label{fig:teaser}
\end{figure}

Frontier models such as GPT-6 Astra, which reconstructs three-dimensional objects by generating CAD code \citep{astraoverview} and builds Blender scenes through computer use \citep{astraarchitecture}, also complete physical manipulation tasks by selecting end-effector targets and gripper commands from visual observations \citep{robocurveastra}: it inspects the scene, chooses a Cartesian pose and gripper state for the robot's inverse kinematics and local controller to realize, and observes the outcome before deciding what to do next. In this setting the model itself is the policy, one motion at a time; GPT-Policy studies it on real robots with in-context demonstrations and human interaction \citep{cheng2026incontextrobotlearningvlm}.

The difficulty is speed. When a manipulation is decomposed into many model-selected motions, each observation--reasoning--action cycle adds waiting time and tokens: in an independent block-placement evaluation, GPT-6 Astra completed 19 of 20 trials at a mean of 2.5 minutes and 2.1k output tokens per run \citep{robocurveastra}, and InSight separates model latency from robot motion in a comparison with CaP-X \citep{insight}. Deployment therefore requires measuring how long a successful interaction takes, alongside whether it succeeds.

Action granularity is one lever. A model asked to specify successive end-effector targets must manage motion details that a local tool can handle; a tool can instead combine several motion phases into one parameterized request, reusable across tasks or written for one. VLA models predict tokenized or continuous low-level actions \citep{openvla,qwenrobotmanip} but generalize poorly to new scene layouts: on RoboDojo, VLA and world action model (WAM) policies keep a median of 12--16\% of their success when the layouts are randomized, whereas GPT-6 Astra acting through RoboDojo's own RoboProbe tool harness keeps 87\% on the official leaderboard \citep{robodojoleaderboard}. Agentic interfaces expose discrete motion commands to a frozen VLM \citep{robodawn,showharness}, and program- and constraint-based systems delegate execution to local controllers \citep{codepolicies,rekep}; the design choice is which decisions stay with the model.

Software development offers a template: with coding agents, developers state what they want, run what the agent writes, and ask for changes, a practice \citet{vibecoding} named vibe coding. We ask whether robot control can work the same way: \emph{can the policy be code that agents write and revise from intent and feedback, so that a frontier model decides which complete motion to run instead of every step, and does this cut end-to-end time and tokens while keeping manipulation reliable?}

Code as Policies showed that a language model can write a robot policy as a program, generated hierarchically and able to call perception \citep{codepolicies}; at run time its code, not the model, decides what runs next, and multi-turn variants regenerate code from execution feedback between runs \citep{capx}. We make code as policy great again with frontier coding agents, keeping the agent in the loop between tool executions (Figure~\ref{fig:teaser}). The robot policy combines agent-written executable tools with feedback-driven tool selection by a frozen foundation model, the execution agent; a programming agent vibe-codes the tools from natural-language intent and execution feedback, starting from reusable tools such as pick-and-place (PnP), PushButton, or Home and writing a task-specific tool when these do not adequately cover a task. The execution agent selects and parameterizes a tool from the current observation and decides again from the feedback after every tool execution; a PnP request covers approach, grasp, transfer, placement, and retreat without a model call for each phase. Unlike casual vibe coding, a tool enters the policy only after it passes validation, and the version used for evaluation is frozen; the two roles separate the cost of writing the policy from the cost of running it. Direct end-effector control also leaves inverse kinematics and local control to the robot; what changes is the granularity and frequency of model decisions: one call executes a complete multi-phase motion, and the model decides the next step from feedback between calls.

The tools run behind \system{}, a universal robot--agent interface that exposes the same tool contract through a GUI and a structured API over one backend that grounds requests, checks planned motion, and executes locally with explicit paths, orientations, timing, and gripper events; humans can steer the programming agent in natural language and operate the same tools through the GUI. Because the tools are code, the policy can also improve without training: guided by failures and natural-language steering, the programming agent revises the tool collection $L$, tool code $C$, and execution strategy $P$ with foundation-model weights fixed; validated versions are retained for later episodes and frozen before evaluation.

Recent systems share parts of this design: Agent as Policy drives a real robot with a coding agent that writes and revises programs from physical feedback, with separate preparation and execution agents \citep{agentaspolicy}; CaP-X adds multi-turn feedback and skill synthesis to Code as Policies \citep{capx}; ENPIRE and ASPIRE refine code policies from rollouts \citep{enpire,aspire}; LATM, CRAFT, and Voyager split tool makers from users or accumulate code skills \citep{latm,craft,voyager}. What distinguishes \system{} is the handoff boundary: a fixed tool collection, a frozen execution agent that decides again after every complete tool execution, and the same tool contract for humans through a GUI, evaluated against direct low-level control and against a program written in advance over the same tools.

Our contributions are fourfold. (1) \emph{Method}: code as policy with frontier agents in every role. A programming agent writes the robot tools; each tool call executes a timed end-effector path that the backend grounds, checks, and runs locally; and a frozen execution agent orchestrates all of these tools through \system{}, a universal robot--agent interface whose GUI and API share one tool contract for humans and agents. (2) \emph{Performance, extensibility, and efficiency}: without robot training data, Claude Opus 5.5 through \system{} reaches a 60.0\% five-task mean over 25 episodes on RoboDojo, close to the best learned policy on the public leaderboard (58.1\%, under its own protocol); in our experiments, a new task called for writing or revising tools rather than collecting demonstrations and retraining; and for three of the four models, deciding one complete motion at a time is faster and uses fewer tokens than step-by-step control by the same model. (3) \emph{Tool evolution in a feedback loop}: the execution agent decides again from the feedback of every call, and the programming agent revises the tool collection and tool code across episodes without weight updates, as in the author-guided extension of a rigid-object PnP tool to cloth folding. (4) \emph{Simulation and real-robot evidence}: on five RoboDojo tasks \citep{robodojo} with four execution agents, \system{} raises success from 18.0\% to 53.0\% over direct fingertip control, deciding after each call reaches 56\% against 24\% for a program written in advance over the same tools, and three of the four agents finish 1.3--1.5$\times$ faster with fewer output tokens; on seven AgileX dual-arm tasks, against published runs of the same model on other robots, \system{} finishes tic-tac-toe 2.7$\times$ faster than GPT-Policy with the same 3/3 successes and block into bowl 2.4$\times$ faster with 4.7$\times$ fewer output tokens than direct end-effector control.

\section{Related Work}
\label{sec:related}

\paragraph{Frontier models as robot decision makers.}
Code as Policies establishes hierarchical program generation, with perception calls inside the program, as an interface to perception and control \citep{codepolicies}; CaP-X extends it with multi-turn interaction, structured execution feedback, and automatic skill synthesis, and exposes a trade-off between the convenience of higher-level primitives and the expressivity of lower-level control \citep{capx}; Agent as Policy drives a physical robot with a coding agent that writes programs and revises them from physical feedback, with a preparation agent for reusable task definitions and an execution agent per trial \citep{agentaspolicy}. VoxPoser composes three-dimensional value maps \citep{voxposer}, MOKA and PIVOT use visual marks and iterative visual prompting \citep{moka,pivot}, ReKep generates constraints that a local optimizer converts into reactive motion \citep{rekep}, RoboStream keeps a frozen VLM in a closed loop through an object-centric scene-graph memory \citep{robostream}, and Inner Monologue feeds environment feedback back into language-model planning \citep{innermonologue}; GPT-Policy takes the step-by-step route with a frontier VLM on real robots \citep{cheng2026incontextrobotlearningvlm}. Persistent improvement outside model weights has precedents in Voyager's code skill library \citep{voyager}, in LATM and CRAFT, which separate tool makers from tool users for language agents \citep{latm,craft}, and in robotics in ASPIRE's validated skill repairs \citep{aspire}, ENPIRE's reset--execute--verify--refine loop on physical rollouts \citep{enpire}, and RHO's search over policy repositories against simulated reward \citep{rho}. \system{} is distinguished by the handoff boundary it evaluates: a fixed collection of agent-written tools, a frozen execution agent that decides again after every complete tool execution, and one tool contract shared by the GUI and the API.

\paragraph{Frontier-model control and action granularity.}
Robocurve evaluates GPT-6 Astra on physical YAM arms through absolute end-effector pose commands, with a 20-call budget and a 25\% speed cap \citep{robocurveastra}, and \citet{astraembodied} compare direct control with hybrid control that accepts or corrects $\pi_{0.5}$ proposals. OpenVLA predicts learned, quantized per-dimension motor tokens \citep{openvla}, Qwen-RobotManip emits continuous action chunks \citep{qwenrobotmanip}, VGGT-DP and Spatial Policy ground learned visuomotor policies in 3D foundation features and spatially conditioned video prediction \citep{vggtdp,spatialpolicy}, and RoboDawn \citep{robodawn} and Show-Harness \citep{showharness} give a frozen VLM a fixed vocabulary of discrete translation, rotation, and gripper commands. \system{} also presents discrete tool calls, but a call can run a reusable tool such as PnP or a task-specific procedure whose internal motion phases execute locally; the distinction is the duration and responsibility assigned to one model-level decision, together with the cost of authoring the tool.

\paragraph{Interfaces and efficiency.}
Trajectory and sketch interfaces condition learned policies or frontier agents on rough trajectories, image-space paths, editable sketches, sparse waypoints, or a browser-based three-dimensional interface \citep{rttrajectory,hamster,actionsketcher,sphinx,via}, and human interfaces range from sketching \citep{sketchinterface} to teleoperation and demonstration capture \citep{aloha,vrteleop,umi}; Show-Harness's GUI lets humans and VLMs issue the same single-step units \citep{showharness}. \system{} likewise shares one action vocabulary between human and model operators, but each entry is a complete, timed tool with explicit metric arguments rather than a fixed increment. Computer-use work treats the observation/action boundary as part of an agent system \citep{webarena,osworld,agents} and motivates measuring efficiency in addition to success \citep{osworldhuman}; in robotics, InSight reduces VLM thinking time through acquired primitives \citep{insight}, and Jev and RoboJEV shorten each decision \citep{jev,robojev}. These systems lower the cost of each decision; \system{} instead lowers the number of decisions per episode.

\section{Code as Policy with the Agent in the Loop}
\label{sec:method}

\subsection{The Policy: Agent-Written Tools and Feedback-Driven Selection}

We write the robot policy as $\pi_{\theta,\varphi}$, where $\theta$ are the fixed weights of the foundation model behind the execution agent and $\varphi=(L,C,P)$ is the external system: the tool collection $L$, the executable tool code $C$, and the execution strategy $P$, meaning the task guidance and call conventions given to the execution agent. At each decision, the execution agent maps the instruction, the current observation, and the execution history to one tool call $(\ell,\xi)$ with $\ell\in L$ and arguments $\xi$. The code $C_\ell$ turns the call into a timed motion that runs locally until it completes or is interrupted, so one policy decision spans a complete multi-phase motion rather than one end-effector target. Adaptation happens in $\varphi$, never in $\theta$. The programming agent proposes $\varphi_{k+1}$ from $\varphi_k$, development feedback, and natural-language intent, runs the candidate, and keeps it only if it passes validation: a vibe-coding loop \citep{vibecoding} in which every accepted version is tested and frozen before evaluation.

\subsection{Interaction Contract}
\begin{figure}[t]
\centering
\includegraphics[width=\textwidth]{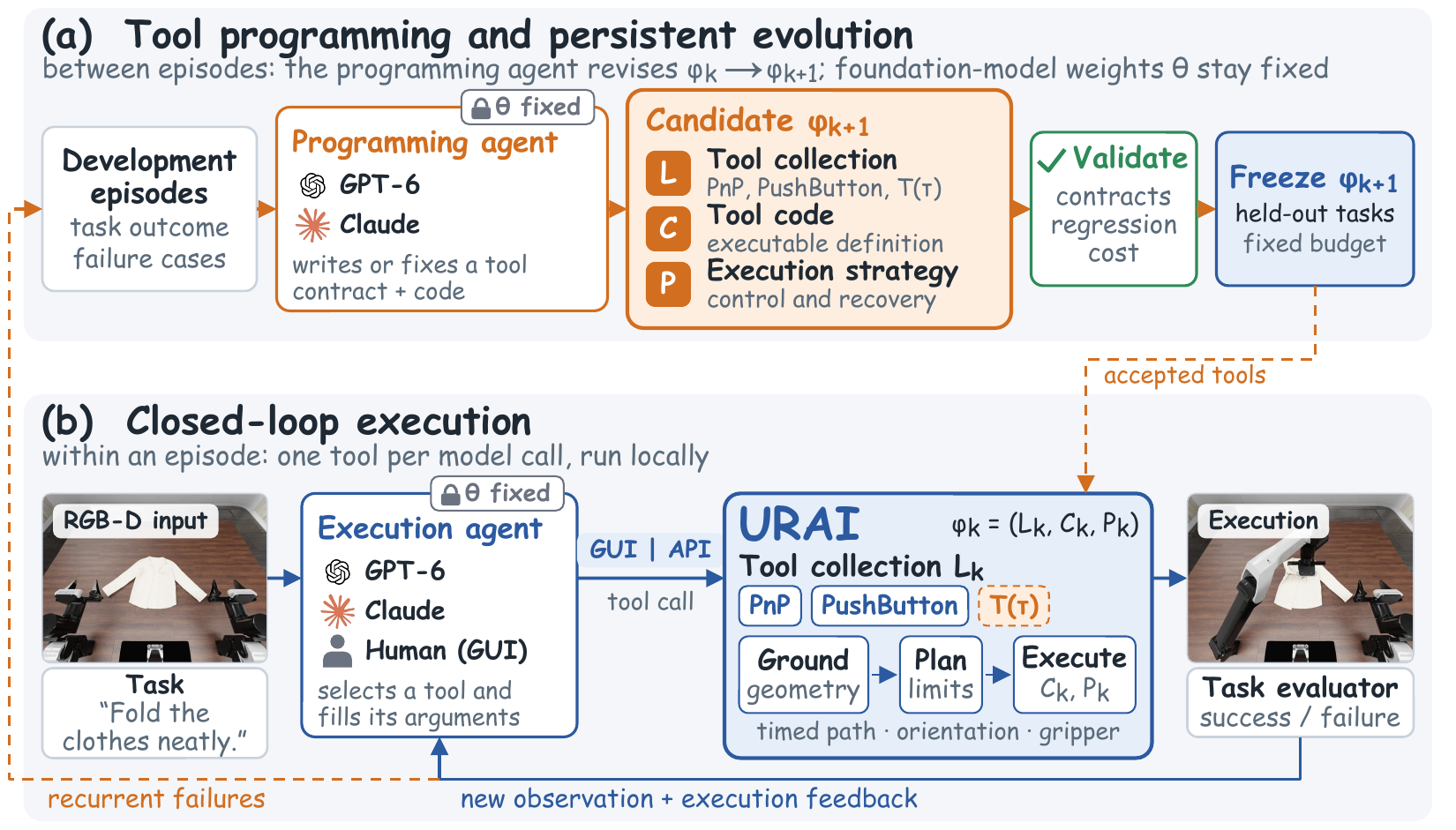}
\caption{\textbf{Tool development and execution in \system{}.} (a) Between episodes, the programming agent revises the tool collection, code, and execution strategy using development feedback. Validated versions are frozen before evaluation; model weights remain fixed. (b) Within an episode, the execution agent selects and parameterizes tools from observations and feedback, while the backend executes each motion locally.}
\label{fig:overview}
\end{figure}

The programming agent receives a task specification, available tool definitions, and development feedback, and can revise a reusable tool or create a task-specific tool, including its code, argument contract, and execution strategy. The execution agent, or a human calling the tools of $L$ through the GUI with the same execution semantics, receives the task instruction, camera and depth-derived observations, robot state, and execution feedback, and invokes an available tool through the GUI or API. The shared backend grounds and validates each request, produces a timed trajectory, and executes it locally; after completion or interruption the execution agent can observe again and issue another request, and recurrent failures can be passed to the programming agent for a versioned tool revision (Figures~\ref{fig:teaser} and~\ref{fig:overview}).

\subsection{A Motion Representation with Explicit Timing}

For an arm, we describe a request as
\begin{equation}
  a = \bigl(p(s), R(s), v(s), E, \mathcal{C}\bigr),\qquad s\in[0,1],
\end{equation}
where $p(s)\in\mathbb{R}^{3}$ is a metric path, $R(s)\in\mathrm{SO}(3)$ is a desired end-effector orientation, $v(s)$ is the requested translational speed profile, $E$ is an ordered set of gripper events and holds, and $\mathcal{C}$ specifies execution settings such as pose tolerances and orientation constraints. Progress $s$ indexes the path; it is not wall-clock time. The prototype accepts sampled control points and bounded mathematical expressions, and interpolates rotations on $\mathrm{SO}(3)$. For a regular translating segment with positive requested speed, the nominal timing relation is
\begin{equation}
  \frac{\mathrm{d}t}{\mathrm{d}s}=\frac{\|p'(s)\|_2}{v(s)}.
  \label{eq:timing}
\end{equation}
This is the classical path--velocity decomposition \citep{bobrow1985,kantzucker1986,toppra}; what the interface adds is that path, speed profile, and gripper events are the arguments of one tool call. Gripper events are tied to motion progress and execution conditions; ordinary placement and dynamic release use different event semantics.

\subsection{Grounding, Editing, and Local Execution}

Points selected on observed surfaces are back-projected from calibrated depth, whereas the height of a free-space path is edited in depth-derived orthographic side views. The GUI adds continuous drawing, arm selection, orientation handles, and a speed-profile editor; the API exposes the corresponding control points, profiles, and motion settings through a bounded arithmetic interpreter, so an agent can specify an entire curve without issuing a mouse action for each point. The backend checks the full trajectory, including the approach from the current pose, against the configured kinematic and geometric constraints and streams references locally at a nominal 50\,Hz on the AgileX backend. The empirical question is whether programmed tools reduce execution-agent interaction costs while maintaining task success.

\section{Experiments}
\label{sec:experiments}
\providecommand{\TBD}{\textit{TBD}}
\providecommand{\NE}{--}
\providecommand{\hd}[1]{\begin{tabular}[c]{@{}c@{}}#1\end{tabular}}

\begin{table}[t]
\centering\scriptsize
\caption{\textbf{Success and execution cost on five RoboDojo tasks.} Upper block: success rates as posted on the public leaderboard \citep{robodojoleaderboard} (official protocol, not paired with ours, no cost data) for the best learned policy of each of the three teams with the highest five-task mean, $\pi_{0.5}$, and GPT-6 Astra through the RoboProbe harness \citep{robodojoastra}. Lower block: five episodes per task (seeds 0--4); success is the official terminal state. Avg.~SR: average success rate, the mean of the five per-task rates (tasks weighted equally). Time and Tokens average all episodes, failures included, and count execution-agent output with reasoning. Program: one program over the same tools, written in advance and run without the model (time includes both phases). Pour balls uses its task-specific pouring tool. Bold: best success per task over all rows (ties included); within the leaderboard block and within each agent (separated by rules), the best Avg.~SR and the lowest time and tokens.}
\label{tab:urai_robodojo_results}
\setlength{\tabcolsep}{2.5pt}\renewcommand{\arraystretch}{0.92}
\begin{tabular*}{\linewidth}{@{\extracolsep{\fill}}lcccccccc@{}}
\toprule
Method & \hd{Pour liquid\\into cup} & \hd{Pour balls\\into vase} & \hd{Fold\\clothes} & \hd{Swap\\blocks\rlap{\smash{$^{\rm a}$}}} & \hd{Stack\\blocks} & \hd{Avg.\\SR $\uparrow$} & \hd{Time\\(min) $\downarrow$} & \hd{Tokens\\(k) $\downarrow$} \\
\midrule
$\pi_{0.5}$ & 28.0 & 13.3 & 38.7 & 0.0 & 13.3 & 18.7 & \NE & \NE \\
GPT-6 Astra, RoboProbe & 8.0 & 4.0 & 72.0 & 0.0 & \textbf{88.0} & 34.4 & \NE & \NE \\
Xiaomi-Robotics-1 & \textbf{64.0} & 46.0 & 73.3 & 0.0 & 21.3 & 40.9 & \NE & \NE \\
Liber-0 Preview & 57.0 & 31.0 & 79.0 & 1.0 & 72.0 & 48.0 & \NE & \NE \\
Simate-beta & 57.3 & 45.3 & \textbf{80.0} & 38.7 & 69.3 & \textbf{58.1} & \NE & \NE \\
\midrule
DeepSeek-V4-Flash, native & 0.0 & 0.0 & 0.0 & 0.0 & 40.0 & 8.0 & 21.4 & 44.3 \\
DeepSeek-V4-Flash + \textbf{\system{}} & 40.0 & \textbf{60.0} & 0.0 & \textbf{100.0} & 40.0 & \textbf{48.0} & \textbf{20.2} & \textbf{40.3} \\
\cmidrule{1-9}
GPT-6 Astra, native & 0.0 & 0.0 & 20.0 & 0.0 & 60.0 & 16.0 & 13.8 & 6.9 \\
GPT-6 Astra, program & 40.0 & 40.0 & 0.0 & 0.0 & 40.0 & 24.0 & \textbf{7.0} & \textbf{4.6} \\
GPT-6 Astra + \textbf{\system{}} & 40.0 & \textbf{60.0} & 0.0 & \textbf{100.0} & 60.0 & \textbf{52.0} & 9.4 & 4.7 \\
\cmidrule{1-9}
Claude Fable 5.1, native & 0.0 & 0.0 & 20.0 & 0.0 & 60.0 & 16.0 & 25.0 & 52.8 \\
Claude Fable 5.1 + \textbf{\system{}} & 20.0 & 40.0 & 20.0 & \textbf{100.0} & 80.0 & \textbf{52.0} & \textbf{17.4} & \textbf{30.4} \\
\cmidrule{1-9}
Claude Opus 5.5, native & 40.0 & 0.0 & 40.0 & 0.0 & 80.0 & 32.0 & 17.4 & 32.3 \\
Claude Opus 5.5, program & 0.0 & 40.0 & 0.0 & 20.0 & 60.0 & 24.0 & 15.1 & 38.0 \\
\textbf{Claude Opus 5.5 + \system{}} & 40.0 & \textbf{60.0} & 40.0 & 80.0 & 80.0 & \textbf{60.0} & \textbf{13.4} & \textbf{20.0} \\
\bottomrule
\end{tabular*}

\vspace{1pt}
\begin{minipage}{\linewidth}\scriptsize
$^{\rm a}$The tools used for swap blocks bundle a task-level place-on-pad procedure with button pressing, so
\system{} and native control differ in tool semantics as well as call granularity; the program
rows share these tools and give the cleaner contrast.
\end{minipage}
\end{table}

We evaluate task success, time, and token cost in simulation across execution agents and interfaces, time and token efficiency on a real robot against published step-by-step control, and author-guided tool revision in a case study. Claude Fable 5.1 in Claude Code serves as the programming agent, and Claude Opus 5.5 in Claude Code for the task-specific pouring tool of pour balls; the execution agent is named per row, and in simulation four different frozen models take that role. Agent conditions use the API tool-call path; a human operator driving the same tools through the GUI succeeds in 23 of 25 simulated episodes.

\subsection{RoboDojo: five-task success rate}
\label{sec:robodojo}

Table~\ref{tab:urai_robodojo_results} compares four frozen execution agents, Claude Fable 5.1 and Claude Opus 5.5 in Claude Code, GPT-6 Astra in the Codex CLI, and DeepSeek-V4-Flash in an open-source coding agent, on multiple RoboDojo tasks \citep{robodojo}: \emph{Pour Liquid Into Cup}, \emph{Pour Balls Into Vase}, \emph{Fold Clothes}, \emph{Swap Blocks}, and \emph{Stack Blocks}. Each agent controls the same simulator either through \system{} tool calls or \emph{natively}, by commanding fingertip targets with a yaw or, for one arm, any end-effector orientation (so pouring is possible natively), and gripper values. Each episode starts a new agent session in a fresh working directory, isolated from memory, ledgers, and other episodes; the agent sees only the camera images and a public depth probe, chooses pixels and arguments itself, and only RoboDojo's own terminal state counts. Each task has one episode per seed 0--4 with the same thirteen reusable tools, frozen before evaluation; for pour balls, a first round with the reusable pouring tool prompted a task-specific pouring tool, developed on separate seeds, with which all its \system{} and program episodes were rerun. The agents mainly used the pouring tool, the PnP tool (stack, and its revised version of Section~\ref{sec:fold_case} for fold), and place-on-pad with button pressing (swap).

\paragraph{Protocol.}
Both conditions use the same task-specific simulator-step budgets (400, 600, 500, 700, and 550, in table order) and a 100-command limit. Agents stop when they judge the task complete or impossible, lack steps for another motion, or reach the command limit, then return both arms home once for official scoring. Wall-clock limits are 45 minutes for \system{} and 60 for native control; agent failures and tasks unfinished at the limit count as failures; one \system{} episode that had already succeeded and returned home when its limit expired counts as a success. Time spans session launch to exit; tokens count execution-agent output, including reasoning where reported.

\paragraph{Results.}
Through \system{}, every agent succeeds more often than under native control (52 vs 16\% for Claude Fable 5.1, 60 vs 32\% for Claude Opus 5.5, 52 vs 16\% for GPT-6 Astra, and 48 vs 8\% for DeepSeek-V4-Flash; 53.0\% against 18.0\% overall), and three of them finish an episode 1.3--1.5$\times$ faster with 1.5--1.7$\times$ fewer output tokens, whereas DeepSeek-V4-Flash saves only about 5\% of its time and 9\% of its tokens. Paired by agent, task, and seed, only \system{} succeeds in 40 of 100 pairs and only native control in 5 (exact two-sided McNemar $p=7.9\times10^{-8}$). The largest single gap is swap blocks, solved in 19 of 20 \system{} episodes and in none of the 20 native ones; without it, \system{} still leads for every agent, 42.5\% against 22.5\% (21 against 5 discordant pairs, $p=0.0025$). A program written in advance over the same tools isolates the agent's decisions between calls: in this control, Claude Opus 5.5 or GPT-6 Astra first inspects the scene with observations, depth probes, and dry runs, then writes one program that calls the tools and branches on their returns, and the program runs without the model. It succeeds in 12 of 50 episodes, against 28 of 50 when the agent decides after each call (17 against 1 discordant pairs, $p=1.4\times10^{-4}$), and in 11 against 19 of 40 without swap blocks (8 against 0, $p=0.0078$). Deciding between calls, not the tool collection alone, therefore accounts for part of the gain. For reference, the best learned policy on the leaderboard averages 58.1\% (63.0\% against Claude Opus 5.5's 55.0\% without swap blocks), and GPT-6 Astra averages 34.4\% through the RoboProbe harness against 52.0\% through \system{}.

\paragraph{Where execution time is spent.}
Table~\ref{tab:robodojo_timing} shows that \system{} replaces many short motion calls with fewer, longer ones: 3.5--10.0 calls per episode lasting 29--72\,s each, versus 21.0--25.6 calls of 19--20\,s natively. For Fable, Opus, and Astra, time outside robot and observation/probe commands falls by about 45\%, 38\%, and 32\%, respectively, consistent with less time spent deciding between motions. DeepSeek shows the opposite pattern: although its motion-call count falls from 25.6 to 10.0, time outside commands rises from 441 to 509\,s, and observations and probes take a larger share of the episode. Coarser motion calls therefore do not ensure lower total cost; the execution agent's observation and decision behavior also matters.

\begin{table}[t]
\centering\scriptsize
\caption{\textbf{Execution-time decomposition on RoboDojo.} Motion and observation/probe columns give mean per-episode shares of wall time; Other is time outside logged commands. All evaluated episodes, including failures, are included. Bold: within each agent, the fewer motion calls and the less time outside logged commands; the other columns describe the composition of an episode.}
\label{tab:robodojo_timing}
\setlength{\tabcolsep}{3pt}
\begin{tabular*}{\linewidth}{@{\extracolsep{\fill}}lcccccc@{}}
\toprule
Condition & \hd{Motion\\calls $\downarrow$} & s/call & \hd{Motion\\(\%)} & \hd{Obs./probe\\(\%)} & \hd{Other\\(\%) $\downarrow$} & \hd{Other\\(s) $\downarrow$} \\
\midrule
Claude Fable 5.1 + \system{} & \textbf{4.3} & 63 & 31 & 31 & \textbf{38} & \textbf{426} \\
Claude Fable 5.1, native & 22.8 & 20 & 34 & 16 & 51 & 772 \\
Claude Opus 5.5 + \system{} & \textbf{3.5} & 72 & 37 & 35 & \textbf{28} & \textbf{250} \\
Claude Opus 5.5, native & 21.8 & 20 & 44 & 18 & 38 & 405 \\
GPT-6 Astra + \system{} & \textbf{5.6} & 44 & 47 & 21 & \textbf{33} & \textbf{188} \\
GPT-6 Astra, native & 21.0 & 19 & 50 & 16 & 34 & 276 \\
DeepSeek-V4-Flash + \system{} & \textbf{10.0} & 29 & 31 & 34 & 36 & 509 \\
DeepSeek-V4-Flash, native & 25.6 & 19 & 40 & 26 & \textbf{34} & \textbf{441} \\
\bottomrule
\end{tabular*}
\par\smallskip
\begin{minipage}{\linewidth}
\footnotesize\raggedright
Seconds per call pools motion calls; other entries are per-episode means.
\end{minipage}
\end{table}

\begin{table}[t]
\centering\small
\caption{\textbf{Real-robot results.} Upper block: tasks with a published GPT-6 Astra reference; \system{} reports three trials per task with mean time, output tokens, and model calls per episode; the references run on other hardware. Lower block: tasks without a reference; progress is the fraction completed (hot-dog: items of four; toss blocks: blocks of six; fold clothes: operator score); success requires the whole task. Bold: the better value within each task where both rows report a comparable measure (success only at equal trial counts; a call budget is not a measured count).}
\label{tab:urai_real_matched}
\setlength{\tabcolsep}{2pt}
\begin{tabular*}{\linewidth}{@{\extracolsep{\fill}}llcccc@{}}
\toprule
Task & Method & Success $\uparrow$ & \hd{Decisions /\\model calls $\downarrow$} & Tokens $\downarrow$ & \hd{Time\\(min) $\downarrow$} \\
\midrule
\multirow{2}{*}{Unscrew bottle cap} & GPT-Policy, published (R3) & \textbf{3/3} & 54.7 & \NE & 17.9 \\
 & GPT-6 Astra + \system{} & \textbf{3/3} & \textbf{24.7} & 1.52k & \textbf{5.1} \\
\midrule
\multirow{2}{*}{Tic-tac-toe} & GPT-Policy, published (R8) & \textbf{3/3} & 69.7 & \NE & 13.6 \\
 & GPT-6 Astra + \system{} & \textbf{3/3} & \textbf{37.0} & 3.26k & \textbf{5.0} \\
\midrule
\multirow{2}{*}{Block into bowl} & Robocurve, published & 19/20 & $\le$20 (budget) & 2.1k & 2.5 \\
 & GPT-6 Astra + \system{} & 3/3 & 6.3 & \textbf{443} & \textbf{1.0} \\
\midrule
\multicolumn{6}{c}{\emph{\system{}-only tasks, ten trials each (success; mean progress); cost over the first three trials}} \\
Hot-dog serving & GPT-6 Astra + \system{} & 10/10 (100\%) & \NE & 2.40k & 7.9 \\
Toss blocks into bowl & GPT-6 Astra + \system{} & 7/10 (91.7\%) & \NE & 4.69k & 8.1 \\
Pour blocks & GPT-6 Astra + \system{} & 10/10 (100\%) & \NE & 2.29k & 3.1 \\
Fold clothes & GPT-6 Astra + \system{} & 8/10 (90.0\%) & \NE & 1.41k & 4.1 \\
\bottomrule
\end{tabular*}
\par\smallskip
\begin{minipage}{\linewidth}
\footnotesize\raggedright
\system{} calls count execution-agent model responses, including observations, polling, and recovery; GPT-Policy counts robot-tool requests; $\le$20 is a call budget. The first bottle-cap trial excludes unmetered diagnosis and repair. \NE{}: not reported.
\end{minipage}
\end{table}

\subsection{Real-robot evaluation}
\label{sec:agilex}
\begin{figure}[t]
\centering
\includegraphics[width=\textwidth]{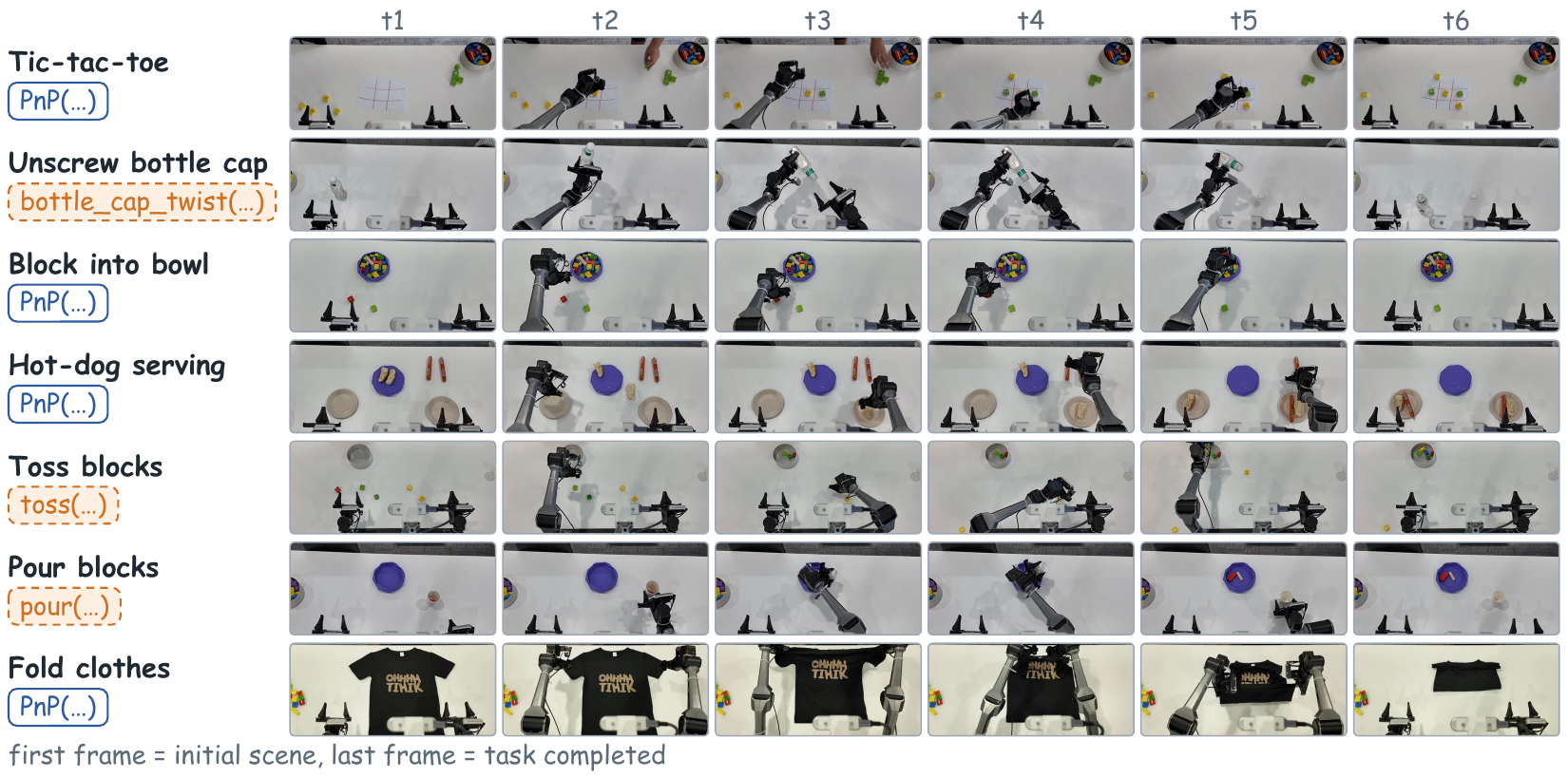}
\caption{Key frames of the seven AgileX tasks under \system{} with GPT-6 Astra, initial scene (t1) to completion (t6); chips name the tool called. Fold clothes is the demonstration-video run; other rows are evaluated trials.}
\label{fig:task_strips}
\end{figure}

The real-robot platform is our AgileX PiPER-X dual-arm system with four RealSense RGB-D cameras. Seven tabletop tasks are run (Figure~\ref{fig:task_strips}), all with GPT-6 Astra as the execution agent calling \system{} tools. Unscrew bottle cap and tic-tac-toe correspond to GPT-Policy's dual-arm tasks R3 (its best condition) and R8 (human interaction, same success criterion) \citep{cheng2026incontextrobotlearningvlm}, and block into bowl repeats Robocurve's task, run there under direct end-effector control \citep{robocurveastra}. Hot-dog serving, tossing blocks into a bowl, pouring blocks, and clothes folding have no published same-task reference and are reported for \system{} alone (lower block).

\paragraph{Measurement protocol.}
The \system{} tasks with published references run at $1\times$ speed; time spans task release to confirmed completion or failure, including robot motion, retries, and waiting for human moves in tic-tac-toe. Output tokens include reasoning; time, tokens, and calls average all three trials, failures included. Robocurve uses YAM arms, a 25\% speed cap, and a 20-call budget over 20 trials per task.

\paragraph{Results.}
All three tasks with a published reference succeed in 3/3 trials (Table~\ref{tab:urai_real_matched}). In tic-tac-toe, the robot lost none of three games against a human (5.0 minutes, 37.0 model calls, and 3.26k output tokens per game), against GPT-Policy's 13.6 minutes and 69.7 decisions with the same model: 2.7$\times$ faster at the same 3/3 success. Unscrew bottle cap took 5.1 minutes and 24.7 calls per trial against R3's 17.9 minutes and 54.7 decisions (3.5$\times$ faster, also 3/3). Block into bowl took 61.3\,s, 6.3 calls, and 443 output tokens per trial against Robocurve's 2.5 minutes, at most 20 calls, and 2.1k tokens (19/20) with the same model under direct end-effector control: 2.4$\times$ faster with 4.7$\times$ fewer tokens. The first trial (71.2\,s, 731 tokens) also wrote the execution script that the other two reused (51.5\,s, 182 tokens; 61.2\,s, 417 tokens); on its own it is still 2.1$\times$ faster with 2.9$\times$ fewer tokens. Among the \system{}-only tasks, run ten times each, hot-dog serving and pour blocks succeed in all ten trials, fold clothes in 8/10 (90.0\% mean progress), and toss blocks into bowl in 7/10 (91.7\%); their token and time columns are means over the first three trials, the only ones with a token log.

\subsection{Tool evolution: revising one tool for fold clothes}
\label{sec:fold_case}
The programming agent, steered by the author in natural language, revised the existing two-point PnP tool of Table~\ref{tab:urai_robodojo_results} seven times (eight versions with the initial one) instead of writing a folding routine. The initial version rejected every click on the shirt as a large rigid object; the revisions add sheet pinching with local thickness checks, synchronized dual-arm transport, outward wrist tilt while carrying, vertical release, and a drop height and clearance taken from depth; for example, replacing the median layer height with its local 0.9 quantile lets the fingers release above an already folded layer. Each version keeps the call signature, ships with unit tests, and has no branch on seed, garment, or task name. The frozen final version is tested by the four frozen execution agents in fresh sessions, choosing pixels themselves: 3 of 20 episodes succeed on seeds 0--4 (Table~\ref{tab:urai_robodojo_results}) and 2 of 20 on seeds 5--9, whereas a human operator choosing pixels through the GUI succeeds in 4 of 5. With this tool, choosing pinch and drop points is the agents' main difficulty: in most failed GPT-6 Astra and Claude Opus 5.5 episodes, the first sleeve pinch sits at the sleeve tip, where the fingers close on nothing.

\section{Limitations and Discussion}

Our measurements cover five simulated and seven real-robot tasks with few episodes per cell; under our protocol, with this tool collection and task mix, all four execution agents have higher aggregate success, and three of them lower execution-phase time and token cost. Pooled over agents, the success gain is significant with and without swap blocks (McNemar $p=7.9\times10^{-8}$ and $p=0.0025$), but five episodes per cell do not establish per-task differences. The program control covers two agents on seeds 0--4; no prompt- or memory-only revision control was run. The real-robot comparisons are against published runs on other hardware with no same-robot step-by-step baseline, so the reported ratios are not causal estimates of a speedup. Tool programming and the author's steering of it cost unmetered time and tokens that may offset the execution-phase savings; the interface is validated on one simulator and one position-controlled dual-arm platform without force regulation, so portability is not established.

\section{Conclusion}

Code as policy with the agent in the loop pairs agent-written tools with feedback-driven tool selection through \system{}, which humans share. On five RoboDojo tasks, four frozen execution agents succeed more often through \system{} than under direct fingertip control, and deciding after each call succeeds more often than a program written in advance over the same tools; three of the four are also faster and use fewer output tokens. Next, we will extend the tool collection to more tasks and robot platforms and meter tool-authoring cost alongside execution cost.

\bibliographystyle{plainnat}
\bibliography{references}
\clearpage
\appendix
\section{Extended Related Work}
\label{app:related}

This appendix gives the full discussion that Section~\ref{sec:related} condenses, preceded by the background on frontier models that the introduction summarizes.

\paragraph{Frontier models in spatial software work and physical manipulation.}
Frontier models are becoming capable of complex spatial work through computer and software tools. OpenAI reports that GPT-6 Astra reconstructs three-dimensional objects from multiview renders by generating CAD code, reaching 95.9\% geometric overlap on BenchCAD \citep{astraoverview}. In a documented design workflow, Astra created an editable house scene in Blender, inspected the application through computer use and rendered previews, and transferred the scene into a walkable Unreal Engine 5 environment; the same account also describes reference-guided garden and other 3D scene construction \citep{astraarchitecture}. These examples demonstrate visual inspection, spatial reasoning, and iterative tool use, although the Blender geometry was produced primarily through scripts rather than mouse-only GUI control. They do not by themselves establish physical manipulation performance.
Frontier models such as GPT-6 Astra have also demonstrated the ability to complete physical manipulation tasks by selecting end-effector targets and gripper commands from visual observations \citep{robocurveastra}. In this form of control, the model inspects the scene, chooses a Cartesian pose and gripper state, executes the command through a robot interface, and observes the outcome before deciding what to do next. The robot's inverse kinematics and local controller realize each command. These results establish that general-purpose models can make useful manipulation decisions on real robots without task-specific training of the frontier model. In this setting the model itself is the policy, one motion at a time; GPT-Policy studies it on real robots with in-context demonstrations and human interaction \citep{cheng2026incontextrobotlearningvlm}.

\paragraph{Frontier models as robot decision makers.}
Code as Policies establishes program generation as an interface to perception and control: the model writes the program first, generating undefined functions hierarchically, and the program can call perception APIs and implement feedback loops and waypoint policies \citep{codepolicies}; at run time its code, not the model, decides what runs next and handles only the cases it checks. VoxPoser uses language models to compose three-dimensional value maps \citep{voxposer}; MOKA uses visual marks to specify affordances and motion \citep{moka}; and PIVOT uses iterative visual prompting for action selection \citep{pivot}. ReKep generates constraints that a local optimizer converts into reactive robot motion \citep{rekep}. These approaches already provide alternatives to predicting each motor command with a frontier model. CaP-X further examines the effects of API abstraction and interaction, adding multi-turn interaction with structured execution feedback and visual differencing, and automatic skill synthesis in CaP-Agent0, and exposing a trade-off between the convenience of higher-level primitives and the expressivity of lower-level control \citep{capx}. Agent as Policy drives a physical robot with a general-purpose coding agent that writes programs, issues motion commands, and revises them from physical feedback; a preparation agent authors reusable task definitions once per task type and an execution agent runs each trial in a fresh session, with model weights fixed \citep{agentaspolicy}. RoboStream is a training-free pipeline that turns each observation into an object-centric scene graph, keeps a short-term causal memory across steps, prompts a frozen VLM for the next manipulation step, and executes the grounded step as a target pose in closed loop \citep{robostream}. These works place the policy in a program whose code makes the run-time decisions, in a composed objective, or in the model's own step-by-step choices; GPT-Policy takes the last route with a frontier VLM on real robots \citep{cheng2026incontextrobotlearningvlm}. We place it in a persistent collection of executable tools that coding agents write, validate, and revise across episodes, in the manner of vibe coding \citep{vibecoding}, while the frontier model at run time chooses and parameterizes complete tools and decides again after every tool execution, rather than leaving later decisions to generated code. \system{} is the interface that makes this possible: an editable motion contract whose visual and programmatic forms share execution semantics, so that humans and agents operate the same tools. Our experiments compare it with direct fingertip-target control in the same simulator, with a Code-as-Policies-style condition in which the agent writes one program over the same tools that then makes the subsequent calls without the model (Appendix~\ref{app:robodojo_tasks}), and with published step-by-step runs on real robots.

\paragraph{Persistent program and policy improvement.}
ASPIRE autonomously writes and repairs code-as-policy robot programs from multimodal execution traces, retains validated repairs in a reusable skill library, and uses evolutionary search to expand the explored task and program space \citep{aspire}. ENPIRE instead provides a real-world experimentation harness in which coding agents operate a repeatable reset--execute--verify--refine loop and improve heuristic or learned policies from physical rollouts \citep{enpire}. RHO has tool-enabled coding agents search, before deployment and with reflective feedback from reward and execution in simulation, for multi-file policy repositories that compose perception, planning, and control primitives; on its manipulation benchmarks the selected repository then runs single-turn, without LLM code edits at deployment, and on RAI's O3DE benchmark, where an LLM stays in the control loop, RHO optimizes that agent's harness of prompts, tools, and control code \citep{rho}. All of its benchmarks are simulated, and it meters the token cost of the search. These systems are close precedents for persistent improvement outside foundation-model weights. Outside robotics, Voyager accumulates a library of code skills for an embodied language agent \citep{voyager}, and LATM and CRAFT separate a tool-making model from a tool-using model \citep{latm,craft}. \system{} focuses on a complementary interface question. Its tools are called at run time by a frozen execution agent that reads feedback between calls; humans operate the same tools through a GUI with the same execution semantics; tool revisions are proposed by a programming agent steered in natural language, validated on development episodes, and kept across episodes, rather than found by reward-driven search; and it is evaluated on a real robot as well as in simulation.

\paragraph{Recent frontier-model control evaluations.}
Robocurve evaluates GPT-6 Astra on physical YAM arms through absolute end-effector pose commands, using three camera views, proprioception, a 20-call budget, and a 25\% speed cap \citep{robocurveastra}. These settings matter when interpreting its time and token measurements. GPT-Policy uses an off-the-shelf VLM with a constrained robot-tool interface and tests human videos, robot demonstrations, goal images, interaction history, and online human input on real robots \citep{cheng2026incontextrobotlearningvlm}. We run two of its stationary tabletop tasks on our AgileX dual-arm platform and exclude its mobile exploration task. \citet{astraembodied} compare direct control with hybrid control that accepts or corrects $\pi_{0.5}$ proposals. Their RoboDojo implementation executes 1--5 steps for direct commands or corrections, versus 1--15 steps when accepting policy actions. Action priors and segment lengths therefore change together; the token savings do not isolate an interface effect. Their reported physical duration excludes model-response latency. RoboDawn gives a frozen VLM a compact interface of discrete translation, rotation, and gripper commands, supplemented by optional in-context demonstrations \citep{robodawn}. It reports RoboDojo \citep{robodojo} and RoboTwin 2.0 evaluations and physical Franka and Piper trials. Its published RoboDojo aggregate covers 42 tasks and cannot be treated as a five-task URAI result. Its released one-demonstration run on our five layouts is a published reference under another protocol, not a control of our study; likewise, GPT-Policy's and Robocurve's physical runs are published references from other hardware, which we compare with \system{} runs on our AgileX platform (Section~\ref{sec:agilex}).

\paragraph{Action tokens and execution granularity.}
OpenVLA casts visuomotor control as language-model prediction of tokenized robot actions: continuous action dimensions are discretized into a finite vocabulary and decoded back to executable values \citep{openvla}. Qwen-RobotManip scales the learned-policy route without action tokens: a Qwen3.5-4B backbone with a flow-matching action expert, pretrained on roughly 38{,}100 hours of robot, human-video, and synthesized data, emits continuous action chunks at 30\,Hz \citep{qwenrobotmanip}. Learned visuomotor policies also draw on spatial priors: VGGT-DP adopts the VGGT 3D foundation model as the visual encoder of a diffusion policy with proprioception-guided visual learning \citep{vggtdp}, and Spatial Policy conditions embodied video generation on an explicit spatial plan, predicts actions with a flow-based module, and replans from spatial reasoning feedback \citep{spatialpolicy}. RoboDawn's textual \texttt{move}, \texttt{rotate}, and \texttt{gripper} commands revive the broader idea of discrete action symbols for a frontier VLM, but they are manually specified semantic primitives rather than OpenVLA's learned, quantized per-dimension motor tokens \citep{robodawn}. Show-Harness takes the same route with a fixed vocabulary of directional moves, axis rotations, and gripper commands; embodiment-specific interpreters map each unit deterministically to one Cartesian pose increment, and the vocabulary is driven zero-shot by Gemini-3.1 Pro or by a LoRA-tuned Qwen3.5-2B on Franka and bimanual AgileX arms \citep{showharness}. \system{} also presents discrete tool calls to the execution agent, but the programming agent can provide both reusable tools such as PnP and task-specific executable procedures. Their internal motion phases execute locally. Thus the distinction is not discrete versus continuous output alone; it is the duration and responsibility assigned to one model-level decision, together with the cost of authoring the tool. Our native condition, in which the same agents command fingertip targets in the same simulator, and our program condition, in which a program written in advance over the same tools makes the calls, place this boundary differently (Appendix~\ref{app:robodojo_tasks}).

\paragraph{Sketches and intermediate representations for manipulation.}
Trajectory sketches are established robot instructions. RT-Trajectory conditions a learned policy on rough trajectories provided by humans, videos, or model-generated specifications \citep{rttrajectory}. Its representation already includes gripper transitions and temporal information; shared human/model input and time-annotated trajectories are therefore precedents, rather than standalone novelty claims for \system{}. HAMSTER trains a high-level VLM to predict image-space paths and a low-level policy to execute them from three-dimensional observations \citep{hamster}. Action-Sketcher incorporates editable visual sketches into a learned reasoning-and-action framework \citep{actionsketcher}. SPHINX combines sparse waypoint actions and dense local control, with a web interface for collecting salient points and waypoints \citep{sphinx}. VIA instead lets off-the-shelf frontier agents control a robot through a browser-based three-dimensional interface \citep{via}; it is an important visual-interface precedent for the native conditions, although its published LIBERO and assembly results are not RoboDojo scores. Our focus is the interface used by off-the-shelf agents and the cost of completing physical tasks through it. We make metric geometry and timing explicit at the command boundary.

\paragraph{Human--robot sketch interfaces and demonstration collection.}
\citet{sketchinterface} study a web-based sketch interface for a mobile manipulator and report lower workload and greater intuitiveness than a conventional axis-control interface in their user study. This work directly motivates evaluating the usability of drawing-based commands. Other interfaces capture detailed human behavior through different hardware: ALOHA supports bimanual teleoperation for imitation learning \citep{aloha}, VR teleoperation provides pose-based demonstrations \citep{vrteleop}, and UMI uses handheld grippers and a carefully designed policy interface to transfer demonstrations to robots \citep{umi}. These systems are relevant precedents for expressivity and data acquisition, but their reported results do not establish a common ranking of interface precision or user friendliness. Show-Harness's GUI Manipulation Interface is closer to our setting: humans and VLMs issue the same single-step semantic units through labeled controls and keystrokes, and each recorded observation--action pair is reusable across embodiments \citep{showharness}. \system{} likewise offers one software action vocabulary to human and model operators, but each entry is a complete, timed tool with explicit metric arguments rather than a fixed increment.

\paragraph{Computer-use interfaces and efficiency.}
WebArena exposes browser observations as screenshots, DOM representations, or accessibility trees, and supports actions grounded by coordinates or element identifiers \citep{webarena}. OSWorld evaluates agents in real desktop applications using execution-based task outcomes \citep{osworld}. Agent S explicitly studies an agent--computer interface within a computer-use framework \citep{agents}. These works motivate treating the observation/action boundary as part of an agent system. The robotics setting adds a crucial distinction: physical geometry must be estimated from sensors, and a command's outcome depends on continuous dynamics and contact. \system{} adopts the interface-design perspective without assuming that physical scenes provide native, error-free accessibility trees. OSWorld-Human motivates measuring inference and interaction efficiency in addition to success \citep{osworldhuman}. In robotics, InSight reports reduced VLM thinking time through acquired primitives \citep{insight}. Jev shortens the decision itself: a non-autoregressive model answers bounded typed questions in one pass at 70--500\,ms end-to-end \citep{jev}, and RoboJEV drives a MuJoCo Franka with two such choices per 1\,cm step, reporting 43/50 physics-checked successes over five tasks with ten seeds each, against 48/50 for a same-scene rule baseline, and no latency or token figures \citep{robojev}. These systems lower the cost of each decision. \system{} instead targets the number of decisions per episode: a programming agent may revise reusable or task-specific tools while a separate execution agent invokes a frozen tool version during evaluation.

\begin{table}[htbp]
\centering\scriptsize
\caption{Closest related systems on the dimensions along which \system{} differs, as described in the respective reports.}
\label{tab:related_diff}
\setlength{\tabcolsep}{3pt}
\renewcommand{\arraystretch}{1.1}
\begin{tabular*}{\linewidth}{@{\extracolsep{\fill}}>{\raggedright\arraybackslash}p{0.14\linewidth}>{\raggedright\arraybackslash}p{0.17\linewidth}>{\raggedright\arraybackslash}p{0.15\linewidth}>{\raggedright\arraybackslash}p{0.16\linewidth}>{\raggedright\arraybackslash}p{0.15\linewidth}>{\raggedright\arraybackslash}p{0.13\linewidth}@{}}
\toprule
System & Who writes the code, and when & Unit of one model decision & Model at run time & Persistence across episodes & Human interface \\
\midrule
Code as Policies \citep{codepolicies} & LM, one program per instruction & the whole program & program decides; perception calls inside & none & -- \\
CaP-X / CaP-Agent0 \citep{capx} & coding agent, one program per turn & one program per turn & re-plans between turns from execution feedback & synthesized skills (CaP-Agent0) & -- \\
Agent as Policy \citep{agentaspolicy} & preparation agent per task type; execution agent per trial & program and motion commands & revises from physical feedback & saved procedures reused & -- \\
ASPIRE \citep{aspire} & agent writes and repairs programs between rollouts & program & program runs & validated skill library & -- \\
ENPIRE \citep{enpire} & coding agent revises the policy between rollouts & policy & policy runs & retained policies & harness \\
LATM / CRAFT \citep{latm,craft} & tool-maker LLM, offline & one tool call by the tool user & tool user decides each call & toolset & -- \\
RoboDawn \citep{robodawn}, Show-Harness \citep{showharness} & fixed command vocabulary, hand-specified & one increment (move, rotate, gripper) & VLM decides every step & in-context demonstrations & Show-Harness GUI shares the units \\
\system{} & programming agent, author-steered, between episodes; frozen before evaluation & one complete, timed tool & execution agent decides after every tool & tool collection, code, and strategy & GUI with the same tool contract \\
\bottomrule
\end{tabular*}
\end{table}

\section{Implementation Details}
\label{app:impl}

This appendix completes the interface description of Section~\ref{sec:method}. Figure~\ref{fig:overview} in Section~\ref{sec:method} details the two loops that Figure~\ref{fig:teaser} summarizes.

\paragraph{Tool calls.}
The execution agent lists the tool collection, reads a tool's description and argument schema (names, ranges, and defaults), and calls one tool at a time with an observation identifier, image pixels chosen from that observation, and its own arguments. An observation identifier stays valid for 120\,s and only until the robot next moves; a call that cites an expired identifier is refused as stale. A tool plans all of its stages before it moves. A dry run returns that plan without motion; a refusal before motion reports which stage failed and why, and costs no simulator steps. After a tool finishes or is interrupted, its reply reports the stages it completed, which is not a declaration that the task is complete; because an identifier expires with the next motion, every call after a motion is grounded in a new observation.

\paragraph{Path inputs and timing rules.}
The prototype accepts sampled control points and bounded mathematical expressions. Rotations are interpolated on $\mathrm{SO}(3)$ rather than by independently interpolating Euler angles. The nominal relation in Equation~\eqref{eq:timing} covers regular translating segments.
Pure rotations, endpoint ramps, and stationary holds have separate timing rules. The implementation also accounts for angular speed and robot-dependent motion limits, so a requested speed specifies the generated reference.

\paragraph{Gripper events.}
Gripper events are associated with progress and execution conditions. For ordinary placement, the sequence is descent to the placement pose, satisfaction of the configured arrival condition, gripper opening, and retreat. Dynamic release uses a different event semantics, with dispatch tied to motion progress and a calibrated release delay.

\paragraph{Visual grounding, GUI, and API.}
A valid depth observation at image pixel $\tilde u=(u,v,1)^\top$ is back-projected using the camera intrinsics $K$ and camera-to-world transform $(R_{wc},t_{wc})$:
\begin{equation}
 p_w=R_{wc}\bigl(D(u,v)K^{-1}\tilde u\bigr)+t_{wc}.
\end{equation}
This expression locates an observed surface point. It does not determine the desired height of an airborne motion at the same pixel. \system{} therefore distinguishes surface-grounded input from free-space path editing. Depth-derived orthographic side views expose height as an editable coordinate, while a three-dimensional preview relates the edited path to the observed scene.

The GUI supports continuous drawing, arm selection, orientation handles, and a speed-profile editor. Explicit gripper events allow path geometry and grasp/release decisions to be edited together. Higher-level drawing modes generate approach, grasp, transport, and placement phases from sparse input, and a human operator can call the tools of the collection with clicked pixels.

The programmatic interface exposes corresponding control points, profiles, motion settings, and tool calls. Expressions are evaluated in a bounded arithmetic interpreter. This route allows an agent to specify an entire curve without issuing a mouse action for each point. Equivalent visual and API requests use the same backend and the same execution semantics; Appendix~\ref{app:robodojo_tasks} lists the images, depth probe, and robot state that each RoboDojo condition receives.

\paragraph{Local execution backend.}
If the requested start differs from the current end-effector pose, the backend constructs an approach before the user-specified motion. Planning checks the full approach and task trajectory against the configured kinematic and geometric constraints. The dual-arm implementation can bring selected arms to their start poses before beginning their task trajectories together. Observations and previews are versioned so that execution uses a plan associated with the current robot state. Changed state requires refreshing and rebuilding the plan.

On the AgileX dual-arm backend, the implementation streams references locally at a nominal 50\,Hz reference-dispatch frequency in joint-position mode. Its model-based checks cover the robot and the calibrated table, and gripper feedback provides an execution signal.

\section{Experimental Setup}
\label{app:setup}

This appendix gives the configuration of the simulated evaluation of Section~\ref{sec:robodojo} and of the real-robot evaluation of Section~\ref{sec:agilex}.

\subsection{RoboDojo}
\label{app:robodojo_tasks}

\paragraph{Tasks.}
The five tasks, with RoboDojo's instruction, its public success criterion, and the episode budget in simulator steps, are \emph{Pour Liquid Into Cup} (``Pour the liquid from the bottle into the cup''; the cup contains the required amount of liquid and the bottle is upright; 400 steps), \emph{Pour Balls Into Vase} (``Pour all the balls from the cup into the vase''; all seven balls inside, cup upright, robot back at origin; 600), \emph{Fold Clothes} (``Fold the clothes neatly''; both sleeves folded inward, hems near the shoulder points, hem line aligned with the shoulder line within the angle threshold, robot back at origin; 500), \emph{Swap Blocks} (``Swap the two blocks using the empty mat, pressing the button after each move''; the two blocks end on each other's original mats after three moves through the empty mat, the button is pressed after each move, robot back at origin; 700), and \emph{Stack Blocks} (``Stack the three blocks with different textures''; all three blocks stacked, robot back at origin; 550). Partial credit that RoboDojo awards, for the sleeves alone or for two of three blocks, is not success.

\paragraph{Seeds.}
Our seed $k$ is layout $k$ of official dataset seed 0. Seeds 0--4, the same five layouts as RoboDawn's released run \citep{robodawn}, are the evaluation seeds of Table~\ref{tab:urai_robodojo_results}. Seeds 5--9 were run separately under the same protocol with the same frozen tools; seeds 5 and 6 also served to diagnose and verify the swap-blocks environment fix described below.

\paragraph{Execution agents.}
Four frozen models act as the execution agent inside their usual coding-agent harnesses: Claude Fable 5.1 (\texttt{claude-fable-5-1}) and Claude Opus 5.5 (\texttt{claude-opus-5-5}) in Claude Code 2.1.281--283 at effort \texttt{xhigh}, restricted to their shell and file-reading tools; GPT-6 Astra (\texttt{gpt-6-astra}) in the Codex CLI 0.154.0 at medium reasoning effort; and DeepSeek-V4-Flash (\texttt{deepseek-v4-flash}) in the open-source pi coding agent 0.73.1 with thinking at \texttt{high}. Each episode runs in a fresh, empty working directory with project memory, hooks, and sub-agents disabled, from a brief that states the task instruction, the public success criterion, the step budget, and the command syntax; nothing about other episodes, seeds, or the success checker is available. The agent reaches the simulator only through one command-line client over ssh, with its own port and container, and every command is logged.

\paragraph{Observations and depth probe.}
An observation returns the 640$\times$480 head-camera image, which looks down on the table from behind the robot, the two wrist-camera images, both arms' state, and the remaining budget; a depth probe returns the world coordinates of a head-camera pixel from public depth. Agents copy the images to their working directory and view them themselves. The same observation and probe commands are available in every condition. Every played simulator step counts against the budget; observations, probes, dry runs, refusals, and failed plans are free.

\paragraph{Conditions.}
Under \system{}, the agent calls the tools of the collection one at a time as described in Appendix~\ref{app:impl}, choosing pixels from a named observation and the remaining arguments itself, and decides the next call after reading each reply. Under \emph{native} control, the agent moves one arm's fingertip centre to a world position, with the tool pointing down at a chosen yaw or at any end-effector orientation given as a quaternion; moves both arms together, with both tools pointing down; or sets a gripper. The same planner refuses infeasible targets at no step cost. The native brief shows the tool-down form, gives the rig facts (table height, arm bases, yaw convention, grasp height), and asks the agent to read the control script's help, which documents the quaternion option; pouring is therefore possible natively. In the \emph{program} condition, run with Claude Opus 5.5 and GPT-6 Astra on seeds 0--4, an episode has two phases. In the planning session, the agent may take observations, view the images, probe depth, read tool schemas, and dry-run tools, but may not move the robot; it then writes one Python program, using only the standard library and NumPy, that calls the same tools through the same client, takes its own observations, and branches on tool replies, the step budget, and depth probes. After the session has ended, the framework runs the program unchanged, without the model; the program may not call models or other programs besides the client, make network requests of its own, or read files outside its folder, and it ends by returning both arms home once.

\paragraph{Stopping, limits, and continuation.}
The agent stops when it judges the task complete or impossible, when the remaining budget cannot cover another motion, or after 100 commands, then returns both arms home once, which records RoboDojo's terminal state; only \texttt{succeeded} counts. In the program condition, the 100-command limit applies to the planning session, and the program applies the other stopping conditions before returning the arms home. Wall-clock limits are 45 minutes for \system{}, 45 minutes for both phases of the program condition together, with at most 30 minutes for the program run, and 60 minutes for the native condition. If a program does not return the arms home, the framework does so before reading the terminal state; a program that is stopped at its limit or crashes counts as a failure. An episode whose model stream ended abnormally was resumed in the same session at most twice, every agent-side failure counts as a failure, and an episode stopped at its wall-clock limit is scored by the terminal state it had reached.

\paragraph{Timing and token accounting.}
Time is the wall clock of the agent's session from launch to exit; in the program condition it spans the planning session and the program run. Model calls, output tokens (reasoning included where the harness reports it), input tokens, and viewed images are read from each harness's own session log; Codex counts a poll of a long-running command as a model call, so call counts are not comparable across harnesses. Means are taken over all episodes, failures included, and all costs are those of the execution agent within evaluation episodes. In the timing decomposition of Table~\ref{tab:robodojo_timing}, motion time includes returning home; seconds per call is pooled motion-command seconds, agent-issued home commands excluded, divided by the motion-call count; percentage columns average per-episode shares, so a ratio of mean times need not equal the mean share; and Other, the time outside logged commands, includes model inference. Codex's polling of long commands locates their end within one polling interval.

\paragraph{Infrastructure failures.}
An episode is voided and rerun with the same agent, condition, task, and seed only when an infrastructure failure unrelated to the agent occurs, for example when the simulator container is stopped by another process on the shared host or the host disk is full before any motion. Agent-side failures, including timeouts and abnormal model streams that exhaust their resumptions, are never voided.

\paragraph{Command audit.}
After each episode, every logged command is checked against the permitted command forms of its condition and the privileged-information rules; in the program condition, the program file and every command the program issued are checked as well. The checks cover commands outside the permitted form, low-level simulator tools called by name, hosts or ports other than the episode's own, motion after returning home and reading the result, real motion during the planning session of the program condition, and reads of the checker source, ledgers, code trees, or other episodes. Flags are reviewed against the command and its output, and an episode with a confirmed violation counts as a failure.

\paragraph{Isolation of evaluation episodes.}
\label{app:isolation}
Every evaluation episode starts a new agent session in a fresh working directory. Claude Code runs with automatic memory and web tools off, hooks and MCP servers disabled, and only its Bash and Read tools enabled; pi runs with \texttt{-{}-no-context-files -{}-no-skills -{}-no-extensions} and a dedicated empty configuration directory; Codex runs non-interactively through \texttt{codex exec}; the global instruction files that Claude Code and Codex load contain no task information. Beyond the task section, the brief gives only the episode's own simulator port and commands; it gives no success-checker source, ledger, development notes, or paths of other episodes. We audited every command of every scored RoboDojo episode, including the commands issued by the programs of the program condition: none accessed memory, a ledger, the checker source, ground-truth keypoints, or files of another episode.

\paragraph{Swap-blocks environment fix.}
On reset, RoboDojo's articulation code rewrites the spring button's drive target in the scene description but does not push it to the physics engine, so the button rests at 79\% of its travel, and the checker's condition that it returns above 90\% after each press can never hold; official success on swap blocks is then impossible under every condition. The backend's environment bootstrap therefore re-applies the declared drive targets after reset for any articulated object, without reference to task, seed, layout, or button position. The change ships with unit tests, touches neither the tools nor the scoring, and was verified on development seeds 5 and 6. Every condition uses the same patched environment, and swap-blocks episodes run before the fix were voided and rerun with it.

\paragraph{Tool collection and freezing.}
The same thirteen reusable tools are exposed in every task under \system{} and in the program condition: pour, pick-and-place, place-in, place-on-pad, push-button, push-object, sweep, swap, insert, intercept, stack-ring, stow, and strike. They were frozen before evaluation in four code trees: pour liquid and stack blocks use the original frozen tree; pour balls uses the original tree with its pour tool replaced by the task-specific pouring tool, frozen before its \system{} and program episodes were rerun; swap blocks uses the original tree with the reset fix above as the only difference; and fold clothes uses its own frozen tree, whose pick-and-place tool is the revised version of Section~\ref{sec:fold_case}. Each episode runs on a copy of its task's tree, whose revision is recorded with the episode, and the execution agent may neither read nor modify the tool code.

\subsection{AgileX real robot}
\label{app:agilex}

\paragraph{Physical platform.}
The real-robot experiments use two AgileX PiPER-X arms (6-DoF, parallel-jaw grippers with a 70\,mm stroke) mounted side by side on a shared base plate at the rear edge of a tabletop, with their bases about 0.59\,m apart, so that both arms reach a common workspace (Figure~\ref{fig:teaser}). Perception uses four Intel RealSense D435 RGB-D cameras: one on each wrist (eye-in-hand) and two fixed scene cameras, one on the central mast looking down on the table and one facing the workspace from the front. The wrist cameras are calibrated eye-in-hand and the overhead camera eye-to-hand against a planar AprilTag board, and the two arm bases are registered to a common frame through the same board; the table plane is fitted from depth at run time and rejected when its residual or tilt exceeds fixed thresholds. Each arm is driven over its own CAN bus at 1\,Mbit/s in joint-position mode; the backend streams joint references from the timed Cartesian path (Section~\ref{sec:method}) and closes the gripper until contact under an effort limit rather than to a fixed width. Perception models run on off-board GPU servers and foundation models are reached through their APIs; the robot host runs only capture, geometry, and control. The scene in Figure~\ref{fig:teaser} shows the props used for these tasks: colored letter and number blocks, cups and a bottle, a cloth, a power strip with push buttons, plates, and a bowl.

\begin{figure}[htbp]
\centering
\includegraphics[width=\textwidth]{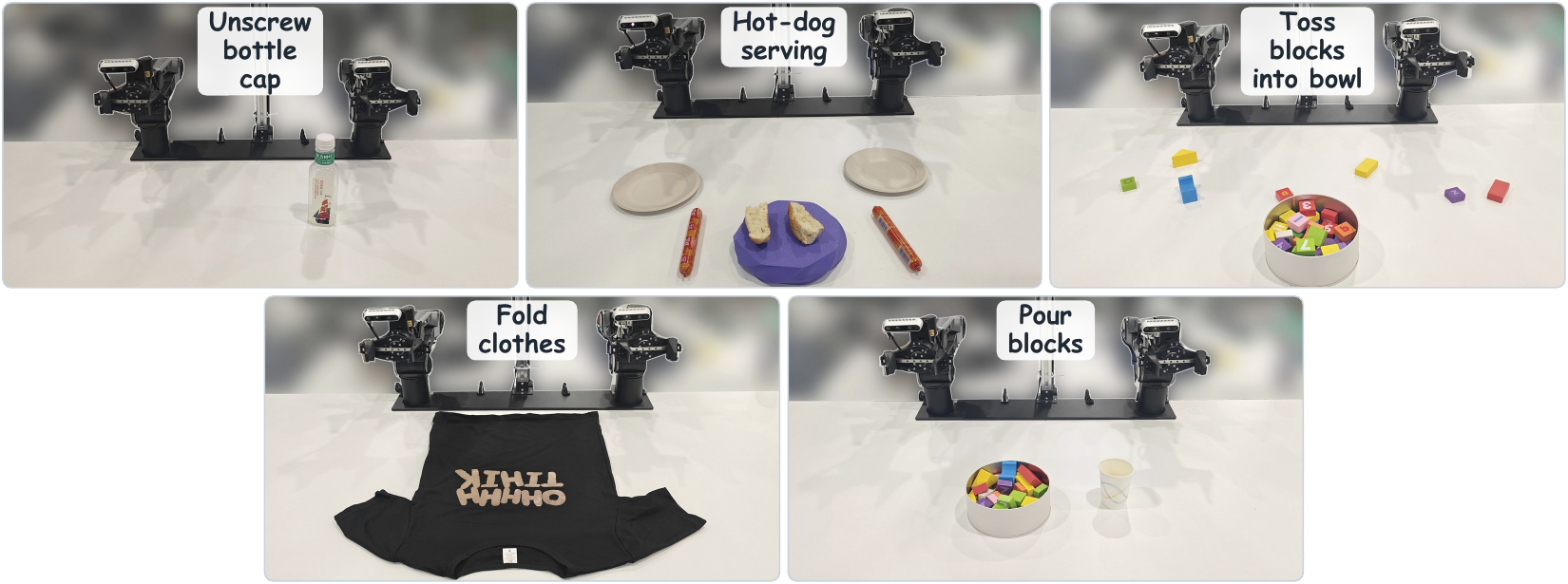}
\caption{Initial scenes of the AgileX dual-arm tasks, seen from the front scene camera and cropped to a common scale: unscrew bottle cap, hot-dog serving, toss blocks into bowl, fold clothes, and pour blocks. Tic-tac-toe and block into bowl are not pictured. Backgrounds are blurred for clarity; the table, arms, and objects are unmodified.}
\label{fig:real_tasks}
\end{figure}

\paragraph{Task set.}
Seven tabletop tasks are run on this platform, all with GPT-6 Astra as the execution agent calling \system{} tools: unscrew bottle cap, tic-tac-toe, block into bowl, hot-dog serving, tossing blocks into a bowl, pouring blocks, and clothes folding; Figure~\ref{fig:real_tasks} shows the initial scenes. The first two correspond to GPT-Policy's stationary dual-arm tasks R3 (unscrew bottle cap, with robot video and action context in its best condition) and R8 (tic-tac-toe, human interaction) \citep{cheng2026incontextrobotlearningvlm}. Block into bowl repeats Robocurve's task, picking up a block from the table and placing it inside a bowl \citep{robocurveastra}. The remaining four tasks have no published same-task reference; clothes folding is the physical counterpart of the simulated task of Section~\ref{sec:fold_case}. GPT-Policy's other stationary tasks and its Movable Exploration task, which requires a mobile base, are not ported.

\paragraph{Task settings and success criteria.}
In unscrew bottle cap, the robot unscrews the cap from a bottle, following R3. In tic-tac-toe, the robot plays against a human opponent, moving first in the first and third games and second in the second, and a game succeeds if the robot wins or draws, R8's criterion. In block into bowl, the robot places one block, white, red, or green in the three trials, inside the bowl. In hot-dog serving, the robot distributes two buns and two sausages so that each of two plates receives one bun and one sausage; progress is the number of items on the plates out of four. In toss blocks into bowl, six blocks lie scattered on the table and all six must end up in the bowl; progress is the number of blocks in the bowl out of six. In pour blocks, the robot pours the blocks held in a cup into a blue container and returns both arms to the home pose. In clothes folding, the robot folds a shirt on the table, and progress is the operator's completion score. In every task, success requires the whole task to be completed.

\paragraph{Trials and speed.}
The three tasks with a published reference, unscrew bottle cap, tic-tac-toe, and block into bowl, are run for three trials each at $1\times$ robot speed. The other four tasks are run for ten trials each; success and progress are computed over all ten trials, and time and tokens over the first three, the only ones with a token log. The logged pour-blocks trials use a robot speed scale of 0.8.

\paragraph{Timing and token accounting.}
\label{app:accounting}
The episode clock starts when the task is released to the execution agent and stops at a confirmed terminal outcome or the common limit. Observation, model inference, network waiting, physical motion, retries, recovery, verification, and, in tic-tac-toe, waiting for the human's moves are included; video acceleration and nominal trajectory duration do not replace this wall-clock measurement. Tokens are execution-agent output tokens with reasoning included. Model calls count one per execution-agent model response within the token-accounting window, including model-initiated observations, polling, status queries, retries, recovery, and replies; loops inside tools are not counted. Token totals include all actual model calls, including repeated context and auxiliary calls, without double-counting cached or reasoning subsets already included in provider usage; missing usage is reported as missing rather than zero. Means are taken over all trials, failures included, and all costs are those of the execution agent. In the first bottle-cap trial, counting starts after a diagnosis-and-repair phase, once the repair was loaded; in block into bowl, the first trial includes generating an execution script that the second and third trials reuse.

\paragraph{Protocol differences from published references.}
GPT-Policy runs its tasks on its own hardware, and its result table does not identify the robot used for each trial. It reports means over three trials and counts decisions, the model's robot-tool requests, which is narrower than our model calls, which also include status queries and replies; it publishes no per-task token counts for R3 or R8. Our R3 counterpart is compared with GPT-Policy's best condition, which gives the same model robot video and action references as context, and our tic-tac-toe keeps R8's human-interaction semantics and success criterion. Robocurve runs GPT-6 Astra on YAM arms through absolute end-effector pose commands, with three camera views, proprioception, medium reasoning effort, a 25\% robot speed cap, and a cap of 20 model calls per run, over 20 trials per task; it publishes per-run output tokens with reasoning included, and its puzzle-insertion task was not run here. Both are therefore published references from other hardware and operating limits, not same-robot controls.

\end{document}